# A Minimax Algorithm faster than NegaScout

Aske Plaat

This report is a description of the MTD(f) algorithm, aimed at practitioners.

This report is made available as a service to the community.

This report is to be available online at http://plaat.nl/publications

MTD(*f*) is a new minimax search algorithm, simpler and more efficient than previous algorithms. In tests with a number of tournament game playing programs for chess, checkers and Othello it performed better, on average, than NegaScout/PVS (the AlphaBeta variant used in practically all good chess, checkers, and Othello programs). One of the strongest chess programs of the moment, MIT's parallel chess program Cilkchess uses MTD(*f*) as its search algorithm, replacing NegaScout, which was used in StarSocrates, the previous version of the program.

MTD(*f*) is only ten lines of code? Here it is (my apologies for mixing C and Pascal; block structure is indicated by indentation only):

**function** MTDF(*root* : node_type; *f* : integer; *d*: integer) : integer;

*g* := *f*;*upperbound* := +INFINITY;*lowerbound* := -INFINITY;**repeat**

**if** *g* == *lowerbound* **then** *beta* := *g* + 1 **else**
*beta* := *g*;*g* := AlphaBetaWithMemory(*root*, *beta* - 1, *beta*, *d*);**if** *g* < *beta* **then** *upperbound* := *g* **else** *lowerbound* := *g*;

**until** *lowerbound* >= *upperbound*;**return** *g*;

The algorithm works by calling AlphaBetaWithMemory a number of times with a search window of zero size. The search works by zooming in on the minimax value. Each AlphaBeta call returns a bound on the minimax value. The bounds are stored in *upperbound* and *lowerbound*, forming an interval around the true minimax value for that search depth. Plus and minus INFINITY is shorthand for values outside the range of leaf values. When both the upper and the lower bound collide, the minimax value is found.

MTD(*f*) gets its efficiency from doing only zero-window alpha-beta searches, and using a "good" bound (variable *beta*) to do those zero-window searches. Conventionally AlphaBeta is called with a wide search window, as in AlphaBeta(*root*, -INFINITY, +INFINITY, *depth*), making sure that the return value lies between the value of alpha and beta. In MTD(*f*) a window of zero size is used, so that on each call AlphaBeta will either fail high or fail low, returning a lower bound or an upper bound on the minimax value, respectively. Zero window calls cause more cutoffs, but return less information - only a bound on the minimax value. To nevertheless find it, MTD(*f*) has to call AlphaBeta a number of times, converging towards it. The overhead of re-exploring parts of the search tree in repeated calls to AlphaBeta disappears when using a version of AlphaBeta that stores and retrieves the nodes it sees in memory.

In order to work, MTD(*f*) needs a "first guess" as to where the minimax value will turn out to be. The better than first guess is, the more efficient the algorithm will be, on average, since the better it is, the less passes the repeat-until loop will have to do to converge on the minimax value. If you feed MTD(*f*) the minimax value to start with, it will only do two passes, the bare minimum: one to find an upper bound of value $x$, and one to find a lower bound of the same value.

Typically, one would call MTD(*f*) in an iterative deepening framework. A natural choice for a first guess is to use the value of the previous iteration, like this:

**function** iterative_deepening(*root* : node_type) : integer;

*firstguess* := 0; **for** $d$ = 1 **to** MAX_SEARCH_DEPTH **do**

*firstguess* := MTDF(*root*, *firstguess*, *d*); **if** times_up() **then** break;

**return** *firstguess*;

In a real program you're not only interested in the value of the minimax tree, but also in the best move that goes with it. In the interest of brevity that is not shown in the above pseudo code.

In case your program has a strong oscillation in the values it finds for odd and even search depths, you might be better off by feeding MTD(*f*) its return value of two plies ago, not one, as the above code does. MTD(*f*) works best with a stable Principal Variation. Although the transposition table greatly reduces the cost of doing a re-search, it is still a good idea to not re-search excessively. As a rule, in the deeper iterations of quiet positions in good programs MTD(*f*) typically performs between 5 and 15 passes before it finds the minimax value.

## Some Background

The name of the algorithm is short for MTD($n$, $f$), which stands for something like Memory-enhanced Test Driver with node $n$ and value $f$. MTD is the name of a group of driver-algorithms that search minimax trees using zero window AlphaBetaWithMemory calls. Judea Pearl has named zero window AlphaBeta calls "Test", in his seminal papers on the Scout algorithm (the basis for Reinefeld's NegaScout). Adding memory to Test makes it possible to use it in re-searches, creating a group of simple yet efficient algorithms.

MTD(*f*) is simple in that it only does zero window AlphaBeta calls, making reasoning about the parts of the tree that get traversed easier than with algorithms that use wide window calls, such as NegaScout and the standard AlphaBeta. Actually, the difficulty in analyzing ordinary AlphaBeta was precisely the reason why Pearl introduced his Test in the first place. The AlphaBeta versions shown on this page can be simplified to use a single input bound, instead of both alpha and beta, since alpha is always one less than beta (null-window).

Especially in a parallel setting the simplicity of MTD(*f*) compared to NegaScout is valuable. Designing and debugging a parallel search routine is a complex affair. MTD(*f*) only needs a zero window search, a Test. Instead of two bounds, MTD(*f*) needs one. In NegaScout, when new values for the search window become available they have to be communicated asynchronously to the child processes; in MTD(*f*) you simply abort an entire subtree when a cutoff happens. Furthermore, the recursive search code does not spawn re-searches anymore. All re-searching is done at the root, where things are simpler than down in the parallel tree. The large body of research on parallelizing AlphaBeta and NegaScout is directly applicable to MTD instances, since they use zero-window AlphaBeta calls to do the tree searching. See for example Bradley Kuszmaul's Jamboree search or Rainer Feldmann's Young Brothers Wait Concept. If your AlphaBeta is parallel, then your MTD(*f*) is parallel.

Incidentally, one of the MTD instances is equivalent to SSS*, George Stockman's best-first minimax algorithm that promised to be more efficient than AlphaBeta. (By equivalent I mean that the two algorithms look different, but search the same nodes.) This SSS*-MTD made the first practical tests of SSS* in full-fledged game playing programs feasible, shedding new and unexpected light, after more than 15 years, on the questions posed in Stockman's 1979 article. See our 1996 article in Artificial Intelligence for more on this subject. (And yes, MTD(*f*) is also better than SSS*, in case you wondered.)

Another instance of the MTD framework is equivalent to the K. Coplan's C* algorithm. Jean-Christophe Weill has published a number of papers on experiments with a negamax version of C*. In MTD terms the idea of C* is to bisect the interval formed by the upper and lower bounds, reducing the number of AlphaBetaWithMemory calls. On the down side, bisection yields a value for the search window, *beta*, that turns out to be not as efficient as MTD(*f*)'s choice. But still, Weill's work indicates that it is worthwhile to experiment with variants on MTD(*f*)'s choice of pivot value.

# AlphaBetaWithMemory

Note that the MTD(*f*) code calls an AlphaBeta version that stores its nodes in memory as it has determined their value, and retrieving their values in a re-search. If AlphaBeta wouldn't do that, then each pass of MTD(*f*) would re-explore most of those nodes. In order for MTD(*f*) to be efficient your AlphaBeta has to store the nodes it has searched. An ordinary tranposition table of reasonable size suffices, as our experiments showed (see [further reading](#)).

To be sure, here's a minimax version of the pseudo code of AlphaBetaWithMemory. The transposition table access code is the same as what is used in most tournament chess, checkers, and Othello programs.

**function** AlphaBetaWithMemory(*n* : node_type; *alpha* , *beta* , *d* : integer) : integer;

**if** retrieve(*n*) == OK **then** /* Transposition table lookup */

**if** *n*.lowerbound >= *beta* **then return** *n*.lowerbound;**if** *n*.upperbound <= *alpha* **then return** *n*.upperbound;*alpha* := max(*alpha*, *n*.lowerbound);*beta* := min(*beta*, *n*.upperbound);

**if** *d* == 0 **then** *g* := evaluate(*n*); /* leaf node */**else if** *n* == MAXNODE **then**

*g* := -INFINITY; *a* := *alpha*; /* save original alpha value */*c* := firstchild(*n*);**while** (*g* < *beta*) **and** (*c* != NOCHILD) **do**

*g* := max(*g*, AlphaBetaWithMemory(*c*, *a*, *beta*, *d* - 1));*a* := max(*a*, *g*);*c* := nextbrother(*c*);

**else** /* n is a MINNODE */

*g* := +INFINITY; *b* := *beta*; /* save original beta value */*c* := firstchild(*n*);

**while** (*g* > *alpha*) **and** (*c* != NOCHILD) **do**

*g* := min(*g*, AlphaBetaWithMemory(*c*, *alpha*, *b*, *d* - 1));*b* := min(*b*, *g*);*c* := nextbrother(*c*);

/* Traditional transposition table storing of bounds *//* Fail low result implies an upper bound */**if** *g* <= *alpha* **then** *n*.upperbound := *g*; store *n*.upperbound;/* Found an accurate minimax value - will not occur if called with zero window */**if** *g* > *alpha* **and** *g* < *beta* **then** *n*.lowerbound := *g*; *n*.upperbound := *g*; store *n*.lowerbound, *n*.upperbound;/* Fail high result implies a lower bound */**if** *g* >= *beta* **then** *n*.lowerbound := *g*; store *n*.lowerbound;**return** *g*;

Transposition table access takes place in the **retrieve** and **store** calls. The lines around retrieve make sure that if a value is present in the table, it is used, instead of continuing the search. The store function is needed to make sure that the table is filled with values as they become available. In a real program, you would also store the best move in the transposition table, and upon retrieving search it first. In the interest of brevity that is not shown in this code.

Text books on Artificial Intelligence typically discuss a version of AlphaBeta that does *not* use memory. Therefore, to avoid any confusion, and even though the use of transposition tables is standard practice in the game playing community, the fact that MTD(*f*) needs a memory-enhanced searcher is stressed here. (Yes, I dislike the name AlphaBetaWithMemory too. Life would be so much simpler if AI text books would discuss practical AlphaBeta versions.) The AlphaBetaWithMemory code is given in the interest of completeness. If you already have a chess program that uses AlphaBeta or NegaScout (minimax or negamax make no difference) and it uses a transposition table, then in all likelihood it will work right away. In none of the programs that I tried did I have to change the existing AlphaBeta code (actually, in all cases a negamax version of NegaScout) to get the transposition table to work properly.

## Implementation Tips

The coarser the grain of eval, the less passes MTD(*f*) has to make to converge to the minimax value. Some programs have a fine grained evaluation function, where positional knowledge can be worth as little as one hundredst of a pawn. Big score swings can become inefficient in for these programs. It may help to dynamically increase the step size: instead of using the previous bound, one can, for example, add an extra few points in the search direction (for failing high, or searching upward, adding the bonus, and for failing low, or searching downward, subtracting the bonus) every two passes or so. (Don Dailey found that a scheme like this works well in a version of Cilkchess.) At the end, if you overshoot the minimax value, you have to make a small search in the opposite direction, using the previous search bound without an extra bonus, to make the final convergence. Also, it can be quite instructive to experiment with different evaluation function grain sizes. Sometimes coarse grain functions work better than fine grain, both for NegaScout and MTD(*f*).

Some programs unroll the search of the root node, for example, to do extra move sorting at the root. In MTD(*f*) you can do this too. In the interest of cleanliness this is not shown in the pseudo code.

Sometimes forward pruning or search extensions that depend on alpha and beta values react in surprising ways to a search that consists of zero window calls only. You may have to do some re-tuning to get it all in synch again. Keep in mind that the size of the search tree is quite sensitive to the value of the search window; a strongly oscillating "first guess" is a bad thing. Also, if it weren't for the transposition table, MTD(*f*)'s re-searches would cause a lot of overhead. Make sure that the transposition table works properly. MTD(*f*) does more re-searches than NegaScout, and tends to take a bigger penalty from a badly functioning transposition table. Consider storing leaf and quiescence nodes in the table. Experiment with different sizes. There is, however, a limit beyond which expanding the table becomes pointless. That limit should be about the same for NegaScout and MTD(*f*).

Some tips to keep in mind when doing experiments to better understand the behavior of your search algorithm: The size of the search tree can differ significantly from position to position. Use a large test set in your experiments. Try different set-ups, such as: without null-move pruning, without extensions, disregard counting of quiescence nodes, different transposition table sizes and storage schemes, disable some parts of the move ordering. As in all debugging, if you want to understand the search better, the idea is to disable as much smarts as possible, to be able to study the behavior of a clean, noise-less, algorithm. (Sure, this will take a lot of time, but it can be quite rewarding to get to understand your program better.)

# Summary

To summarize, the core ideas of MTD(*f*) are:

- The narrower the AlphaBeta window, the more cutoffs you get, the more efficient the search is. Hence MTD(*f*) uses only search windows of zero size.
- Zero window AlphaBeta calls return bounds. At the root of the tree the return bounds are stored in *upperbound* (after AlphaBeta "failed low") and *lowerbound* (after AlphaBeta "failed high"). The bounds delimit the range of possible values for the minimax value. Each time MTD(*f*) calls AlphaBeta it gets a value back that narrows the range, and the algorithm is one step closer to hitting the minimax value.
- Storing nodes in memory gets rid of the overhead inherent in multiple re-searches. A transposition table of sufficient size does the job.
- It is more efficient to start the search close to its goal. Therefore MTD(*f*) needs (and can make use of) a good first guess.

# NegaScout

Here's a negamax version of the NegaScout code from [Alexander Reinefeld's homepage](), the creator of the algorithm. Note that you have to add the transposition table access code in the appropriate places yourself. (It's a "NegaScout**Without**Memory".)

```
int NegaScout ( int p, alpha, beta );
{                           /* compute minimax value of position p */
   int a, b, t, i;

   determine successors p_1,...,p_w of p;
   if ( w = 0 )
      return ( Evaluate(p) );                        /* leaf node */
   a = alpha;
   b = beta;
   for ( i = 1; i <= w; i++ ) {
      t = -NegaScout ( p_i, -b, -a );
      if (t > a) && (t < beta) && (i > 1) && (d < maxdepth-1)
         a = -NegaScout ( p_i, -beta, -t );       /* re-search */
      a = max( a, t );
      if ( a >= beta )
         return ( a );                              /* cut-off */
      b = a + 1;                            /* set new null window */
   }
   return ( a );
}
```

# References

Here are some publications that describe MTD(*f*), and why it works:

- The MTD(*f*) algorithm and the experiments that showed that it worked in practice were described in award winning University of Alberta technical report, Aske Plaat, Jonathan Schaeffer, Wim Pijls, Arie de Bruin: A New Paradigm for Minimax Search, University of Alberta Technical Report 94-18, (a.k.a. Erasmus University report EUR-CS-95-03), December 1994.
- It was subsequently published in a short paper at IJCAI-95 Aske Plaat, Jonathan Schaeffer, Wim Pijls, Arie de Bruin: Best-First Fixed-Depth Game-Tree Search in Practice, In: Proceedings IJCAI'95, Montreal, Canada, August 1995.
- MTD(*f*) was described in a longer article in the November 1996 issue of the Journal Artificial Intelligence: Aske Plaat, Jonathan Schaeffer, Wim Pijls, Arie de Bruin: Best-First Fixed-Depth Minimax Algorithms, Artificial Intelligence, Issue 1-2, November 1996, which also deals with the relation of other MTD instances to SSS*-like best-first minimax algorithms.
- The algorithm is one of the topics of my 1996 PhD thesis Aske Plaat: Research Re: search & Re-search, PhD thesis, Erasmus University Rotterdam, June 1996
- The work on MTD(f) grew out of research into the SSS* algorithm. A report on this work is Aske Plaat, Jonathan Schaeffer, Wim Pijls and Arie de Bruin: SSS* = Alpha-Beta + TT, Technical Report 94-17, Department of Computing Science, University of Alberta, December 1994.
- If you are interested in further work to on searching smaller trees, you may also want to have a look at: Aske Plaat, Jonathan Schaeffer, Wim Pijls and Arie de Bruin: Nearly Optimal Minimax Tree Search?, Technical Report 94-19, Department of Computing Science, University of Alberta, December 1994.
- The work on the SSS* algorithm has led to a deeper understanding of minimax in terms of Solution Trees. A more theoretical report on this perspective is: Wim Pijls, Arie de Bruin, and Aske Plaat: **A theory of game trees, based on solution trees**, Technical Report 96-06, Erasmus University Rotterdam, 1996.

These papers, and more, with their full reference information, can be found on this list of publications. The work on MTD(*f*) was part of an extraordinary fruitful and pleasant cooperation with Jonathan Schaeffer, Wim Pijls, and Arie de Bruin, which I gratefully acknowledge. The project also benefited from the kind support of Jaap van den Herik and Henri Bal.

# Acknowledgements

Thanks to Peter Kouwenhoven and Yngvi Bjornsson for suggestions for improvements of this page. Thanks to Toru Yamazato for pointing out a bug in the AlphaBetaWithMemory pseudo code.